\documentclass[runningheads]{llncs}
\usepackage{url}
\usepackage{xcolor}
\usepackage{graphicx}
\usepackage{amsmath}
\usepackage{amsfonts}
\usepackage{multirow}
\usepackage{booktabs}
\usepackage{xcolor}
\usepackage[export]{adjustbox}
\usepackage{algorithm}
\usepackage{algpseudocode}
\usepackage{tabularx}
\usepackage{ bbold }
\usepackage{hyperref}

\newcommand{\etal}{\textit{et al.}}
\newcommand{\ie}{\textit{i.e.}}

\newcommand{\mysubtitle}[1]{\noindent{\bf #1}}

\begin{document}
\title{ Date Estimation in the Wild of Scanned Historical Photos: An Image Retrieval Approach}
\titlerunning{Date Estimation in the Wild: An Image Retrieval Approach}
\author{Adrià Molina\orcidID{0000-0003-0167-8756} \and Pau Riba\orcidID{0000-0002-4710-0864} \and Lluis Gomez\orcidID{0000-0003-1408-9803} \and Oriol Ramos-Terrades\orcidID{0000-0002-3333-8812} \and Josep Lladós\orcidID{0000-0002-4533-4739}}
%
%\author{Annonymous authors}
%\authorrunning{Annonym Author et al.}
%\institute{Annonymous institute}
\authorrunning{A. Molina et al.}
\institute{Computer Vision Center and Computer Science Department, \\ Universitat Aut\`onoma de Barcelona, Catalunya\\
\email{\{priba,lgomez,oriolrt,josep\}@cvc.uab.cat},\\ \email {adria.molinar@e-campus.uab.cat}}
\maketitle              % typeset the header of the contribution
\begin{abstract}
    This paper presents a novel method for date estimation of historical photographs from archival sources. The main contribution is to formulate the date estimation as a retrieval task, where given a query, the retrieved images are ranked in terms of the estimated date similarity. The closer are their embedded representations the closer are their dates. Contrary to the traditional models that design a neural network that learns a classifier or a regressor, we propose a learning objective based on the nDCG ranking metric. We have experimentally evaluated the performance of the method in two different tasks: date estimation  and date-sensitive image retrieval, using the DEW public database, overcoming the baseline methods.

\keywords{Date Estimation \and Historical Photographs \and Image Retrieval \and Ranking Loss \and Smooth-nDCG.}
\end{abstract}

\section{Introduction}
    \label{sec:Introduction}
Historical archives and libraries contain a large variability of document sources that reflect the memory of the past. The recognition of the scanned images of these documents allows to reconstruct the history. A particular type of archival data are historical photographs which are full of evidence that tells us the story of that snapshot in time. One just needs to pay attention to the subtle cues that are found in different objects that appear in the scene: the clothes that people wear, their haircut styles, the overall environment, the tools and machinery, the natural landscape, etc. All of these visual features are important cues for estimating its creation date. Apart from that, texture and color features might also be of great help to accurately estimate of image creation date since photographic techniques have evolved throughout history and have imprinted a specific date fingerprint on them.

Date estimation of cultural heritage photographic assets is a complex task that is usually performed by experts (e.g.  archivists or genealogists) that exploit their expert knowledge about all the features mentioned above to provide precise date estimations for undated photographs. But their manual labor is costly and time consuming, and automatic image date estimation models are of great interest for dealing with large scale archive processing with minimal human intervention.

Most approaches in date estimation for historical images try to directly compute the estimation through classification or regression~\cite{ginosar2015century,muller2017picture,salem2016analyzing}. As alternative of these classical approaches, in this paper we present a method for date estimation of historical photographs in a retrieval scenario. Thus, the date estimation of photographs is incorporated in the ranked results for a given query image. This allows to predict the date of an image contextualized regarding the other photographs of the collection. In the worst case, when the exact date is not exactly estimated, the user can obtain a relative ordering (one photograph is older than another one), which is useful in archival tasks of annotating document sources. The proposed model for historical photograph retrieval is based in a novel ranking loss function \emph{smooth-nDCG} based on the Normalized Discounted Cumulative Gain ranking metric, which is able to train our system according to a known relevance feedback; in our case the distance in years between images.

The main idea in our approach relies on optimizing rankings such that the closer is image's date to the query's date for a certain photograph the higher will be ranked. 
When receiving an unknown image as query the method computes the distances towards a support dataset, consisting of a collection of images with known dates. The date is estimated assuming that the highest ranked images are images from the same date.

In contrast to the literature reviewed in Section \ref{sec:sota}, our method allows not only to predict but to rank images given a query. This means that considering an image from a certain year our system is capable of retrieving a list of images ordered by time proximity. This may be useful for many applications and systems that rely on retrieving information from a large amount of data.

The rest of the paper is organized as follows: In Section \ref{sec:sota} we present the most relevant state of the art related to our work. The key components of the proposed model are described in Section \ref{sec:learning_obj}, where we describe the learning objectives considered in the training algorithms, and in Section \ref{sec:arch}, where we outline the architecture and training process of the model. Section \ref{sec:experiments} provides the experimental evaluation and discussion. Finally, Section \ref{sec:conclusions} draws the conclusions.

\section{Related Work}
    \label{sec:sota}

The problem of automatic estimation of the creation date of historical photographs is receiving increased attention by the computer vision and digital humanities research communities. The first works go back to Schindler~\etal~\cite{schindler2007inferring,schindler2010probabilistic}, where the objective was to automatically construct time-varying 3D models of a city from a large collection of historical images of the same scene over time. The process begins by performing feature detection and matching on a set of input photographs, followed by structure from motion (SFM) to recover 3D points and camera poses. The temporal ordering task was formulated as a constraint satisfaction problem (CSP) based on the visibility of structural elements (buildings) in each image. 

More related with the task addressed in our work, Palermo~\etal~\cite{palermo2012dating} explored automatic date estimation of historical color images based on the evolution of color imaging processes over time. Their work was formulated as a five-way decade classification problem, training a set of one-vs-one linear support vector machines (SVMs) and using hand-crafted color based features (e.g. color co-occurrence histograms, conditional probability of saturation given hue, hue and color histograms, etc.). Their models were validated on a small-scale dataset consisting of 225 and 50 images per decade for training and testing, respectively.

In a similar work, Fernando~\etal~\cite{fernando2014color} proposed two new hand-crafted color-based features that were designed to leverage the discriminative properties of specific photo acquisition devices: color derivatives and color angles. Using linear SVMs they outperformed the results of~\cite{palermo2012dating} as well as a baseline of deep convolutional activation features~\cite{donahue2014decaf}. On the other hand, martin~\etal~\cite{martin2014dating} also showed that the results of~\cite{palermo2012dating} could be slightly improved by replacing the one-vs-one classification strategy by an ordinal classification framework~\cite{frank2001simple} in which relative errors between classes (decades) are adequately taken into account. 

More recently, Müller~\etal~\cite{muller2017picture} have brought up to date the task of image date estimation in the wild by contributing a large-scale, publicly available dataset and providing baseline results with a state-of-the-art deep learning architecture for visual recognition~\cite{43022}. Their dataset, Date Estimation in the Wild (DEW), contains more than one million Flickr\footnote{\url{https://www.flickr.com/}} images captured in the period from 1930 to 1999, and covering a broad range of domains, e.g., city scenes, family photos, nature, and historical events. 

Beyond the problem of unconstrained date estimation of photographs, which is the focus of this paper, there are other related works that have explored the automatic prediction of creation dates of certain objects \cite{lee2015linking,vittayakorn2017made}, or of some specific types of photographs such as yearbook portraits~\cite{ginosar2015century,salem2016analyzing}.

%\cite{ginosar2015century} 2015 deep learning, VGG pre-trained on the ILSVRC dataset,  large-scale historical image dataset of yearbook portraits, which we have made publicly available~\footnote{https://people.eecs.berkeley.edu/~shiry/projects/yearbooks/yearbooks.html}

%\cite{salem2016analyzing} 2016 \cite{ginosar2015century} concurrent with deep learning , CaffeNet pre-trained on imagenet, 65-way classification output Yearbook-Face Dataset 719 229 images of face patches, and 565 069 images of torso patches. dataset available upon request~\footnote{http://cs.uky.edu/~salem/face2year/}

%\cite{vittayakorn2017made} 2017 deep learning methods for estimating when objects (not photos) were made.

%\cite{lee2015linking} 2015 find visual patterns in the architecture of buildings, relate them to certain time periods, and show how they can be used to date buildings. 

In this work, contrary to all previously published methods, we approach the problem of date estimation from an image retrieval perspective. We follow the work of Brown~\etal~\cite{brown2020smooth} towards differentiable loss functions for information retrieval. More precisely, our model learns to estimate the date of an image by minimizing the the Normalized Discounted Cumulative Gain.

In all our experiments we use Müller \etal's Date Estimation in the Wild~\cite{muller2017picture} dataset. We also share with ~\cite{muller2017picture} the use of state of the art convolutional neural networks in contrast to classic machine learning and computer vision approaches, such as~\cite{fernando2014color}.

\section{Learning Objectives}
    \label{sec:learning_obj}
As mentioned in Section \ref{sec:Introduction} our retrieval system relies on a neural network \cite{he2016deep} trained to minimize a differentiable ranking function \cite{brown2020smooth}.
Traditionally, information retrieval evaluation has been dominated by the mean Average Precision (mAP) metric. In contrast, we will be using the Normalized Discounted Cumulative Gain (nDCG). The problem with mAP is that it fails in measuring ranking success when the ground truth relevance of the ranked items is not binary.

In the task of date estimation the ground truth data is numerical and ordinal (e.g. a set of years in the range $1930$ to $1999$), thus given a query the metric should not punish equally for ranking as top result a 1-year difference image than a 20-years difference one. This problem is solved in the nDCG metric by using a relevance score that measures how relevant is a certain sample to our query. This allows us to not only to deal with date estimation retrieval but to explicitly declare what we consider a good ranking.

%Before discussing how did we manage to optimize the rankings so closer years get to the top we present the following formulations.

In this section we derive the formulation of the smooth-nDCG loss function that we will use to optimize our date estimation model.

\mysubtitle{Mean Average Precision}: %As we already exposed, success in a information retrieval system has been traditionally measured by the \emph{mean Average Precision} (mAP), a classic information retrieval metric~\cite{rusinol2009performance}. 
First, let us define \emph{Average Precision} (AP) for a given query $q$ as

\begin{equation}\label{eq:ap}
    \operatorname{AP}_q = \frac{1}{|\mathcal{P}_q|} \sum_{n=1}^{|\Omega_q|}P@n \times r(n),
\end{equation}

\noindent where \(P@n\) is the precision at \(n\) and \(r(n)\) is a binary function on the relevance of the \textit{n}-th item in the returned ranked list, $\mathcal{P}_q$ is the set of all relevant objects with regard the query $q$ and \(\Omega_q\) is the set of retrieved elements from the dataset. Then, the mAP is defined as:
\begin{equation}
    \operatorname{mAP} = \frac{1}{Q}\sum_{q=1}^{Q} \operatorname{AP}_q,
\end{equation}
where $Q$ is the number of queries.

\mysubtitle{Normalized Discounted Cumulative Gain}: In information retrieval, the \emph{normalized Discounted Cumulative Gain} (nDCG) is used to measure the performance on such scenarios where instead of a binary relevance function, we have a graded relevance scale. The main idea is that highly relevant elements appearing lower in the retrieval list should be penalized. In the opposite way to mAP, elements can be relevant despite not being categorically correct with respect to the query. The \emph{Discounted Cumulative Gain} (DCG) for a query \(q\) is defined as

\begin{equation}\label{eq:dcg}
    \operatorname{DCG}_q = \sum_{n=1}^{|\Omega_q|} \frac{r(n)}{\log_2(n+1)},
\end{equation}%
where \(r(n)\) is a graded function on the relevance of the \textit{n}-th item in the returned ranked list and $\Omega_q$ is the set of retrieved elements as defined above. In order to allow a fair comparison among different queries that may have a different sum of relevance scores, a normalized version was proposed and defined as

\begin{equation}\label{eq:ndcg}
    \operatorname{nDCG}_q = \frac{\operatorname{DCG}_q}{\operatorname{IDCG}_q},
\end{equation}

\noindent where $\operatorname{IDCG}_q$ is the ideal discounted cumulative gain, \ie~assuming a perfect ranking according to the relevance function. It is formally defined as
\begin{equation}
    \operatorname{IDCG}_q = \sum_{n=1}^{|\Lambda_q|} \frac{r(n)}{\log_2(n+1)}
    \label{eq:idcg}
\end{equation}
\noindent where \(\Lambda_q\) is the ordered set according the relevance function.

As we have exposed, we want to optimize our model at the retrieval list level, but the two classical retrieval evaluation metrics defined above are not differentiable. In the following we define two `smooth' functions for ranking optimization that are inspired in the $\operatorname{mAP}$ and $\operatorname{nDCG}$ metrics respectively.

\mysubtitle{Ranking Function} (R). Following the formulation introduced in~\cite{qin2010general}, these two information retrieval metrics can be reformulated by means of the following ranking function,

\begin{equation}
    \mathcal{R}(i, \mathcal{C}) = 1 + \sum_{j=1}^{|\mathcal{C}|} \mathbb{1}\{(s_i- s_j)<0\},
    \label{eq:ranking}
\end{equation}

\noindent where $\mathcal{C}$ is any set (such as $\Omega_q$ or $\mathcal{P}_q$),  \(\mathbb{1}\{\cdot\}\) is the Indicator function, and $s_i$ is the similarity between the $i$-th element and the query according to $S$. In this work we use the cosine similarity as $S$ but other similarities such as inverse euclidean distance should work as well. Let us then define cosine similarity as:

\begin{equation}
S(v_q, v_i) = \frac{v_q \cdot v_i}{\|v_q\|\|v_i\|}.
\end{equation}

\begin{figure}[t]
    \centering
    \includegraphics[scale = 0.31]{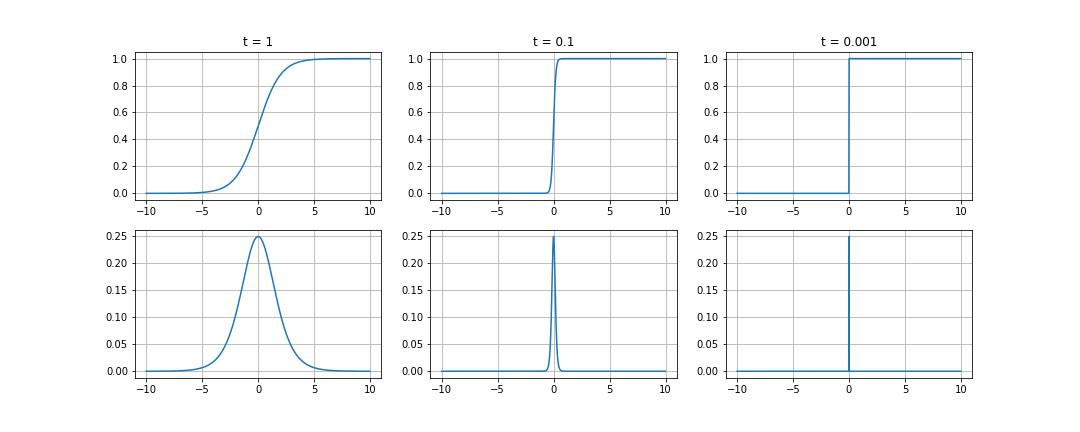}
    \caption{Shape of the smooth indicator function (top) and its derivative (bottom) for different values of the temperature: 1 (left), 0.1 (middle), 0.001 (right).}
    \label{fig:tauComparision}
\end{figure}

Still, because of the need of a indicator function \(\mathbb{1}\{\cdot\}\), with this formulation we are not able to optimize following the gradient based optimization methods and we require of a differentiable indicator function. Even though several approximations exists, in this work we followed the one proposed by~Quin~\etal~\cite{qin2010general}. Thus, we make use of the sigmoid function 
\begin{equation}
    \mathcal{G}(x;\tau) = \frac{1}{1+e^\frac{-x}{\tau}}.
    \label{eq:sigmoid}
\end{equation}

\noindent where $\tau$ is the temperature of the sigmoid function. As illustrated in Figure~\ref{fig:tauComparision} the smaller is $\tau$ the less smooth will be the function. For small values the gradient is worse defined but the approximation is better since the sigmoid becomes closer to the binary step function. 

\mysubtitle{Smooth-AP}: 
%The smoothed approximation of AP, namely Smooth-AP, proposed by Brown~\etal~\cite{brown2020smooth}, has shown a huge success on image retrieval. 
Following~\cite{brown2020smooth} and replacing the term $P@n \times r(n)$ in Eq.~\ref{eq:ap} by the Ranking Function introduced in Eq.~\ref{eq:ranking}, the $\operatorname{AP}$ equation becomes

\begin{equation}
    \operatorname{AP}_q = \frac{1}{|\mathcal{P}_q|} \sum_{i\in \mathcal{P}_q} \frac{1 + \sum_{j\in \mathcal{P}_q, j \neq i} \mathbb{1}\{D_{ij}<0\}}{1+\sum_{j\in \Omega_q, j\neq i} \mathbb{1}\{D_{ij}<0\} }.
\end{equation}

\noindent where we use $D_{ij} = s_i - s_j$ for a more compact notation. Finally, replacing the indicator function by the sigmoid function (Eq. ~\ref{eq:sigmoid}) we obtain a smooth approximation of $\operatorname{AP}$

\begin{equation}
    \operatorname{AP}_q \approx \frac{1}{|\mathcal{P}_q|} \sum_{i\in \mathcal{P}_q} \frac{1 + \sum_{j\in \mathcal{P}_q, j \neq i} \mathcal{G}(D_{ij};\tau)}{1+\sum_{j\in \Omega_q, j \neq i} \mathcal{G}(D_{ij};\tau) }.
\end{equation}%
Averaging this approximation for all the queries in a given batch, we can define our loss as

\begin{equation}
    \mathcal{L}_{AP} = 1 - \frac{1}{Q}\sum_{i=1}^Q AP_q,
    \label{eq:loss_ap}
\end{equation}%
where \(Q\) is the number of queries.

\mysubtitle{Smooth-nDCG}: Following the same idea as above,  we replace the $n$-th position in Eq.~\ref{eq:dcg} by the ranking function, since it defines the position that the $i$-th element of the retrieved set, and the DCG metric is expressed as %
\begin{equation}
    \operatorname{DCG}_q = \sum_{i\in \Omega_q} \frac{r(i)}{\log_2\left(2 + \sum_{j\in \Omega_q, j \neq i} \mathbb{1}\{D_{ij}<0\}\right)},
\end{equation} %
where $r(i)$ is the same graded function used in Eq.~\ref{eq:dcg} but evaluated at element $i$. Therefore, the corresponding smooth approximation is %
\begin{equation}
    \operatorname{DCG}_q \approx \sum_{i\in \Omega_q} \frac{r(i)}{\log_2\left(2 + \sum_{j\in \Omega_q, j \neq i} \mathcal{G}(D_{ij};\tau)\right) }
\end{equation} %
when replacing the indicator function by the sigmoid one.

The smooth-nDCG is then defined by replacing the original $\operatorname{DCG}_q$ by its smooth approximation in Eq.~\ref{eq:ndcg} and the loss $\mathcal{L}_{nDCG}$ is defined as %
%
%Note that this loss is then normalized with the IDCG described in Equation~\ref{eq:idcg}.
%
\begin{equation}
    \mathcal{L}_{nDCG} = 1 - \frac{1}{Q}\sum_{i=1}^Q nDCG_q,
    \label{eq:loss_dcg}
\end{equation}%
where \(Q\) is the number of queries.

%Once defined those functions, the training algorithm Alg.~\ref{alg:train}~mainly relies on using them in the output space for optimizing the CNN so it minimizes the smooth-NDCG loss function Eq. \ref{eq:loss_dcg} and, consequently, maximizes the NDCG metric in our rankings, meaning that images will be closer in the output space as they are closer in the ground truth space.

\section{Training process}
    \label{sec:arch}
\begin{figure}[t]
    \centering
    \includegraphics[scale = 0.60]{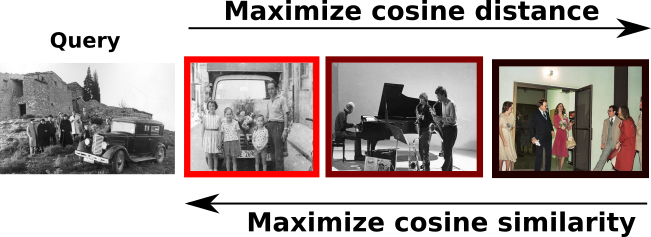}
    \caption{Objective function example, images with closer ground truth should maximize the similarity in the output space.}
    \label{fig:retrieval_ilustration}
\end{figure}

The main contribution of this paper relies on using the smooth-nDCG loss defined in the previous section for optimizing a model that learns to rank elements with non-binary relevance score. Specifically, our model learns to project images into an embedding space by minimizing the smooth-nDCG loss function. In all our experiments we use a Resnet-101 \cite{he2016deep} convolutional neural network pretrained on the ImageNet dataset. 
%As we previously mentioned, we propose a clustering approach where we rank images from a certain embedding. The main contribution this paper relies on is all about using smooth-NDCG for optimizing non-binary labeled information structures where a category may has an ordinal sense. Thus, this embedding is generated through minimizing the smooth-NDCG loss function using a pretrained on ImageNet Resnet-101 \cite{he2016deep}.  With this approach we will be rating not only how good our model is predicting, but how capable is it at looking for images semantically similar between each others. This means the model won't work as a predictor itself but a retrieval system.
%This should generate a vector space such that
As illustrated in Figure ~\ref{fig:retrieval_ilustration} the idea is that in the learned embedding space the cosine similarity between image projections is proportional to the actual distance in years. We will discuss in Section \ref{sec:experiments} how teaching the system to rank images according to the ground truth criteria leads the embedding space to follow a certain organization (illustrated in Figure \ref{fig:DistanceMatrix}).

In order to obtain such embedded representation of the photographs, we present the training algorithm (see Algorithm~\ref{alg:train})~ that mainly relies on using the equations presented in the previous Section \ref{sec:learning_obj} in the output space for optimizing the CNN so it minimizes the smooth-nDCG loss function (see Eq.~\ref{eq:loss_dcg}) and, consequently, maximizes the nDCG metric in our rankings, meaning that images will be closer in the output space as they are closer in the ground truth space.

\begin{algorithm}[H]
\hspace*{\algorithmicindent} \textbf{Input:} Input data $\{\mathcal{X}, \mathcal{Y}\}$; CNN $f$; distance function D; max training iterations $T$ \\ 
\hspace*{\algorithmicindent} \textbf{Output: } Network parameters $w$
\begin{algorithmic}[1]
    \Repeat
    \State Process images to output embedding $h \leftarrow f_w(\{x_i\}_{i=1}^{N_{batch}})$
    \State Get Distance matrix from embeddings, all vs all rankings $M \leftarrow D(h)$
    \State Calculate relevance from training set $Y$ Eq. \ref{eq:relevance}
    \State Using the relevance score, $\mathcal{L} \leftarrow$ Eq. ~\ref{eq:loss_dcg}
    \State $w \leftarrow w - \Gamma(\nabla_{w} \mathcal{L})$ \label{alg:line:H}
    \Until{Max training iterations $T$}
\end{algorithmic}
\caption{Training algorithm for the proposed model.} \label{alg:train}
\end{algorithm}

%The main novelty we propose is the retrieval approach. Then, we suggest a baseline where Resnet-101 should transform our input images to a embedding where distances between are proportional to label distances Fig. \ref{fig:retrieval_ilustration}. This should lead the rankings to get that ordinal sense we don't get from average precision optimization.
\begin{figure}[ht]
    \centering
    \includegraphics[scale = 0.6]{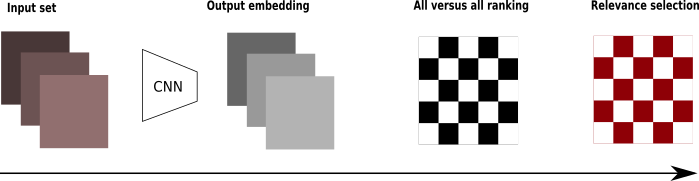}
    \caption{Proposed baseline. Once ranked the images in the nDCG space we'll be computing smooth-nDCG in order to back-propagate the loss through the CNN.}
    \label{fig:baseline}
\end{figure}

As shown in Figure \ref{fig:baseline}, once we process the images through the CNN, we compute the rankings for each one of the batch images. We use cosine similarity as distance function, however it could be replaced by any distance function model over the vector space. Since nDCG requires a relevance score for each sample, the design of the relevance function is an important part of the proposed method. A possible relevance function could be just the inverse distance in years as defined in Eq.~\ref{eq:relevanceNaive}.

% Additionally, we decided to consider only the closest decade in the relevance function (illustrated in Figure \ref{fig:relevances}), so every image further than 10 years is considered as relevant as any other. Eq.~\ref{eq:relevance} is the relevance function  used in our model (see Eq.~\ref{eq:loss_dcg}) but other relevance settings could be, like for example, using exponential  functions (see Eq.~\ref{eq:relevanceLog}) in order to exponentially punish more the images that should not be as far as they are.

\begin{figure}
    \centering
    \includegraphics[scale=0.75]{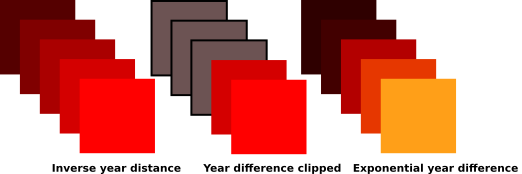}
    \caption{Different examples of relevance functions for ground truth data.}
    \label{fig:relevances}
\end{figure}

Additionally, we decided to consider only the closest decade in the relevance function (illustrated in Figure \ref{fig:relevances}), so every image further than 10 years is considered as relevant as any other. Eq.~\ref{eq:relevance} is the relevance function  used in our model (see Eq.~\ref{eq:loss_dcg}) but other relevance settings could be, like Eq.~\ref{eq:relevanceLog} in order to exponentially punish more the images that should not be as far as they are. %
\begin{align}
    r(n;\gamma) & = \max(0,\gamma - | y_q - y_n |) \label{eq:relevance}\\
    r(n) &= \frac{1}{1+ | y_q - y_n  |} \label{eq:relevanceNaive}\\
    r(n) &= e^{ \frac{1}{1+| y_q - y_n|}} \label{eq:relevanceLog}
\end{align}%
where $y_q$ and $y_n\in\mathcal{Y}$ are the dates of the query and the $n$-th image in the training set, respectively; and $\gamma$, in Eq.~\ref{eq:relevance}, is an hyper-parameter that has experimentally set to 10.

Once trained, minimizing the retrieval loss function, we can predict the year for an unknown image through k-Nearest Neighbors (k-NN) to a support set. %Since using the whole dataset for kNN search is not efficient in terms of computation time % 

\begin{figure}
    \centering
    \includegraphics[scale = 0.4]{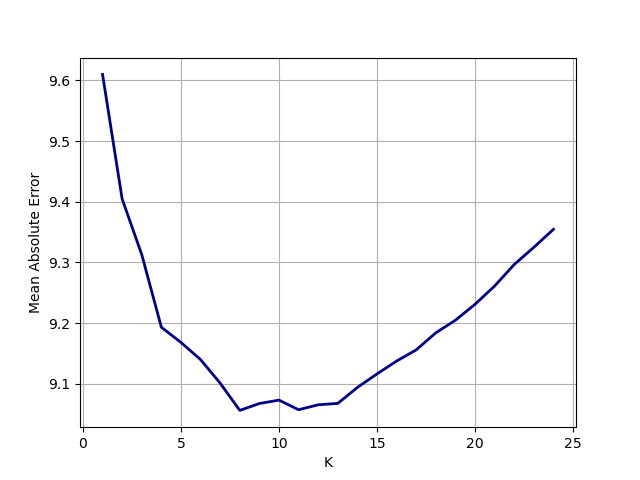}
    \caption{Mean Absolute Error (MAE) on prediction by k-Nearest Neighbors with respect to each k using a randomly selected support set of 128 images}
    \label{fig:kNN}
\end{figure}

In one hand, we can use a set of \textit{N} train images as support set; each batch is randomly selected for each prediction. Note that, since prediction relies on the \textit{k}-th most similar images, the bigger the support set the better should perform the prediction.
In Figure~\ref{fig:kNN} we show the date estimation mean absolute error of our method on the DEW training set depending on the \textit{k} parameter for the k-NN search. Hence, we would be using 10-Nearest Neighbors in all our experiments for the prediction task.

\begin{figure}
    \centering
    \includegraphics[scale = 0.4]{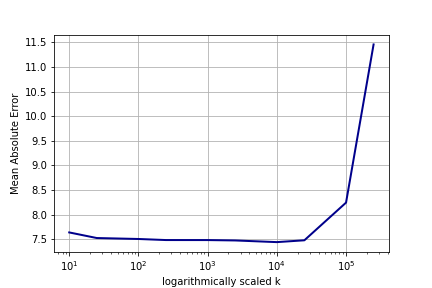}
    \caption{Mean Absolute Error with different $k$ for k-NN using the entire train set as support set. Note that the horizontal axis grows logarithmically.}
    \label{fig:KNN-1M}
\end{figure}

Additionaly, we can compute the retrievals within the entire training set with Approximate Nearest Neighbours\footnote{Using public Python library ANNOY for Approximate Nearest Neighbours \href{https://github.com/spotify/annoy}{\url{https://github.com/spotify/annoy}}}, which allow us to compute the retrievals efficiently. As it's shown in Figure \ref{fig:KNN-1M}, the possibility of using the whole train set makes the $k$ parameter more likely to be huge. We don't observe a notable divergence in the results until using a an enormous $k$ parameter, which may indicate quite good result in terms of retrieval since worst responses are at the bottom of the dataset.  

\section{Experiments}
    \label{sec:experiments}
In this section we evaluate the performance of the proposed method in two different tasks: date estimation and date-sensitive image retrieval. In all our experiments we use the DEW~\cite{muller2017picture} dataset, which is composed of more than one million images captured in the period from $1930$ to $1999$. The test set is balanced across years and contains $1120$ images, $16$ images per year. The temperature parameter, $\tau$, in Eq.~\ref{eq:sigmoid}, is set to 0.01. 

For the task of date estimation we use the Mean Absolute Error (MAE) as the standard evaluation metric. Given a set of $N$ images and their corresponding ground truth date annotations (years) $y=\{y_1, y_2, \dots, y_N\}$, the $\operatorname{MAE}$ for a set of predictions $y'=\{y_1', y_2', \dots, y_N'\}$ is calculated as follows: %
\begin{equation}
    \operatorname{MAE} = \frac{1}{N} \sum_{i}^N{| y_i - y_i' |}
\end{equation}

For the image retrieval task we use the mAP and nDCG metrics defined in Section~\ref{sec:learning_obj}. In Table \ref{tab:baselines} we present a comparison of our methods with the state of the art in the task of date estimation and with a visual similarity baseline, that ranks images using the euclidean distance between features extracted from a ResNet CNN~\cite{he2016deep} pre-trained on ImageNet. As we will discuss in the conclusions section, our approach is not directly comparable to Müller \etal's \cite{muller2017picture}. On one hand, we are not trying to directly improve the prediction task, but a ranking one; this means that our model has to learn semantics from the images such that it learns a continuous ordination. On the other hand, the way we are estimating image's date is by using a ground truth dataset or support dataset. Nevertheless, we observe that our model isn't too far with respect to the state of the art prediction model. We consider this a good result since estimating the exact year of a image is not the main purpose of the smooth-nDCG loss function. 
Note that in Table \ref{tab:baselines} we mention a "cleaned test set". This is because we observed the presence of certain placeholder images indicating they are no longer available in Flickr. We computed the MAE for the weights given by Müller \etal \cite{muller2017picture} \footnote{This is an extended version of Müller et al. ~\cite{muller2017picture} made publibly available by the same authors at \url{https://github.com/TIB-Visual-Analytics/DEW-Model}} for a proper comparison.  The presence of this kind of images may have a low impact while evaluating a classification or regression, but they can have a higher impact on predicting with k-NN, since all the retrieved images will be placeholders as well independently of their associated years.

%\begin{figure}
    %\centering
    %\includegraphics[scale=0.4]{images/placeholder_flickr.jpg}
    %\caption{Example of useless placeholder image that may bias the %evaluated metrics.}
    %\label{fig:placeholder}
%\end{figure}

We present as well a result of predicting the year by weighted k-NN where the predicted year will be the sum of the $k$ neighbours multiplied by their similarities to the query.
%\label{eq:weightedKnn}
%    y_q' =\frac{1}{\sum_{i}^k{s_{qi}}}\sum_{i}^k{s_{qi}  y_i} 
%\end{equation}

\begin{table}[t]
\caption{Mean absolute error comparison of our baseline model on the test set of the Date Estimation in the Wild (DEW) dataset.}
\label{tab:baselines}
\begin{tabularx}{\columnwidth}{Xccc}
\toprule
Baseline & MAE & mAP & nDCG\\   
\midrule
Müller et al.~\cite{muller2017picture} GoogLeNet (regression) & 7.5   & - & -\\
Müller et al.~\cite{muller2017picture} GoogLeNet (classification) & 7.3 & - & - \\
\midrule
Müller et al. ~\cite{muller2017picture}  ResNet50 (classification) (cleaned test set) & 7.12 & - & -\\
\midrule
Visual similarity baseline & 12.4 & - & 0.69 \\
\midrule
Smooth-nDCG 256 images support set & 8.44 & 0.12 & 0.72 \\
Smooth-nDCG 1M images support set & 7.48 & - & 0.75\\
Smooth-nDCG 1M images support set (weighted kNN) & 7.52 & - &  0.75\\
\bottomrule
\end{tabularx}
\end{table}

Since our model is trained for a retrieval task, we provide further qualitative examples of good and bad rankings. This exemplifies how our loss function is not mean to predict years, but finding good rankings according to query's year. Note we computed nDCG using as relevance the closeness in years between images' ground truth clipped at 10 years (see Eq.~\ref{eq:relevance}). Despite it may seem nDCG is so high and mAP so low; there are many subtle bias in this appearance. Since the ground truth can be interpreted as a continuous value (such as time) there are not many matches in any retrieval of  \textit{N} random images for a certain query regarding to query's year compared with how many negative samples are there. Then, mAP hardly punishes results in this situation. In the opposite way, the same thing happens to nDCG. Since nDCG has this `continuous' or `scale' sense, any retrieval will have some good samples that satisfies the relevance function (see Eq.~\ref{eq:relevance}), being the results way less punished if there aren't many exact responses in the retrieval. Briefly, we could say nDCG brings us a more balanced positive/negative ratio than mAP for randomly selected sets to rank (which is absolutely more realistic than forcing it to be balanced). Additionally, nDCG allows us to approach retrievals that are not based on obtaining a category but a wide number of them, or even retrieving non-categorical information such as date estimation.

\begin{figure}[t]
    \centering
    \includegraphics[scale=0.5]{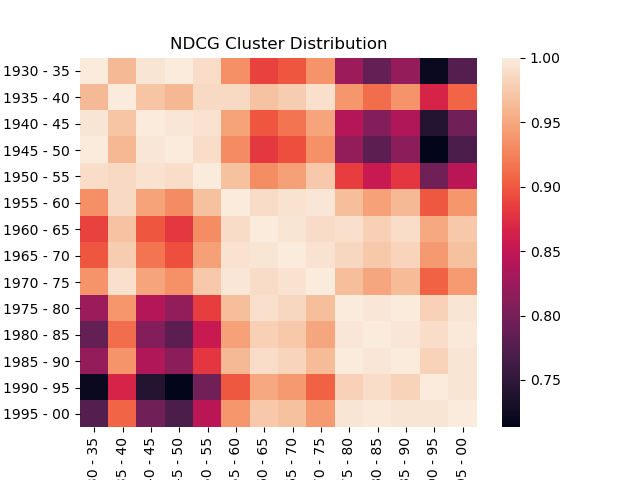}
    \caption{Clusters average similarity for each 5 year interval. We observe a certain visible pattern in the diagonal; the closer the years the closer the clusters.}
    \label{fig:DistanceMatrix}
\end{figure}

Additionally, as we observe in Figure \ref{fig:DistanceMatrix}, we compute the average cosine similarity between clusters from different groups of years. This shows us how our output space is distributed with respect to the ground truth data. As we expected; clusters for closer years are the most similar between themselves. This is because ranking is nothing but taking the closer points to our query; by forcing our system to take those points that satisfies the relevance function (see Eq.~\ref{eq:relevance}), points in the embedding space will have to be organized such that distances are proportional to ground truth's distances.

\begin{figure*}
    \centering
    \setlength\tabcolsep{3pt}
    
    \includegraphics[width=\linewidth]{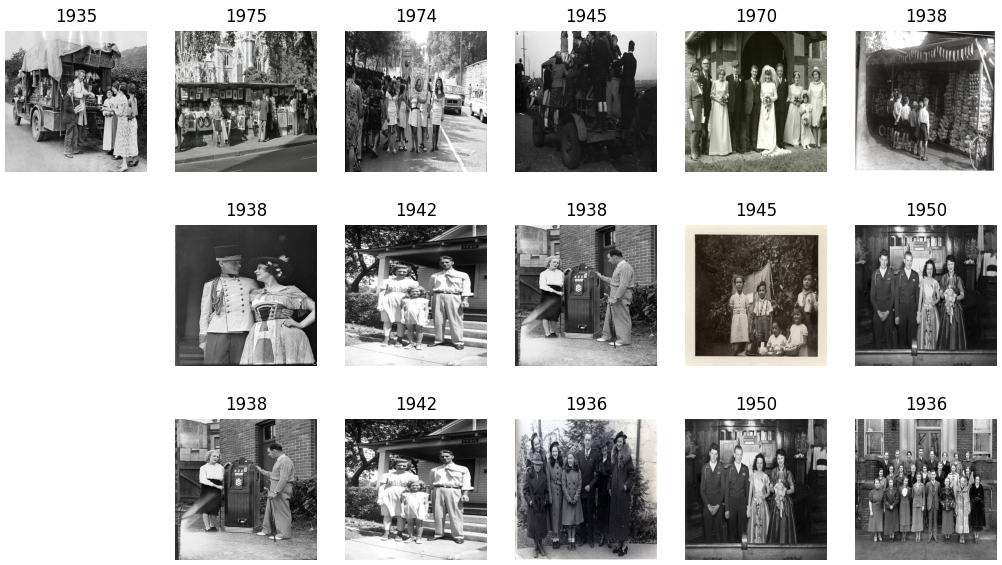}
    
    \caption{Qualitative results for a retrieval system that combines our date estimation model and a visual similarity model. The top-left image is the query, in each row we show the top-5 retrieved images using a visual similarity model (top), our model (middle), and the combination of the two rankings (bottom).}
    \label{fig:combined_ranking}
\end{figure*}

Finally, in Figure \ref{fig:combined_ranking}, we compare the top-5 retrieved images for a given query with our model and with the visual similarity baseline model, that ranks images using the distance between features extracted from a ResNet CNN~\cite{he2016deep} pre-trained on ImageNet. We appreciate that the visual similarity model retrieves images that are visually similar to the query but belong to a totally different date. On the contrary, our model performs better for the same query in terms of retrieving similar dates although for some of the top ranked images the objects layout is quite different to the query. Additionally, we present the results with a mixed model that re-ranks images with the mean ranking between both methods for each image in the test set; in this way we obtain a ranking with visually similar images from the same date as the query.

% \begin{equation}\label{eq:relevance}
% rel = \max(10 - | y_{\mathrm{query}} - y_{\mathrm{gt}} |, ~0)
% \end{equation}

% \begin{equation}\label{eq:relevanceNaive}
% rel = \frac{1}{1+ | y_{\mathrm{query}} - y_{\mathrm{gt}}  |}
% \end{equation}

% \begin{equation}\label{eq:relevanceLog}
% rel = e^{ \frac{1}{1+| y_{\mathrm{query}} - y_{\mathrm{gt}}|}}
% \end{equation}

\begin{figure*}
    \centering
    \setlength\tabcolsep{3pt}
    
    \includegraphics[width=\linewidth]{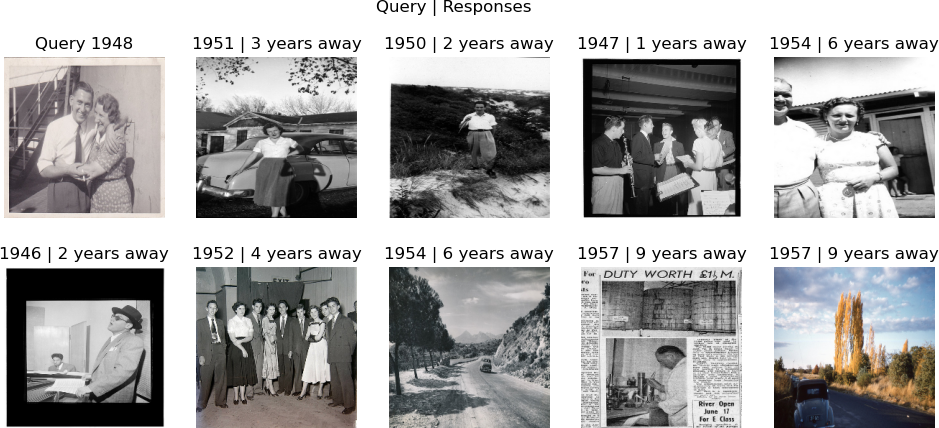}
    
    \caption{Qualitative results for our date estimation model. We show a set of images from the DEW test set and indicate their ground truth year annotation and distance to the query.}
    \label{fig:qualitative}
\end{figure*}

\section{Conclusions}
    \label{sec:conclusions}
In this paper we have proposed a method for estimating the date of historic photographs. The main novelty of the presented work is that we do not formulate the solution to the problem as a prediction task, but as an information retrieval one. Hence metrics such as mean absolute error do not have to be considered as important as retrieval metrics. Even so, we managed to build an application-level model that can estimate the date from the output embedding space with a decent error according to previous baselines \cite{muller2017picture}. Since our prediction method is essentially a clustering or k-Nearest Neighbors method we are using a known support set, so predictions relies on already labeled data. 

However, considering the output embedding space Figure \ref{fig:DistanceMatrix} we conclude that smooth-nDCG function works pretty well for ranking large-scale image data with deeper ground truth structures; unlike smooth-AP \cite{brown2020smooth}, smooth-nDCG cares about the whole sorted batch retrieved, not only how many good images are, but how are the bad ones distributed along the ranking. This allowed us to retrieval ordered information (such as years), or tree-structured information where different categories can be closer or further depending on how the tree is like. 

\begin{figure}[t]
    \centering
    \includegraphics[width=\linewidth]{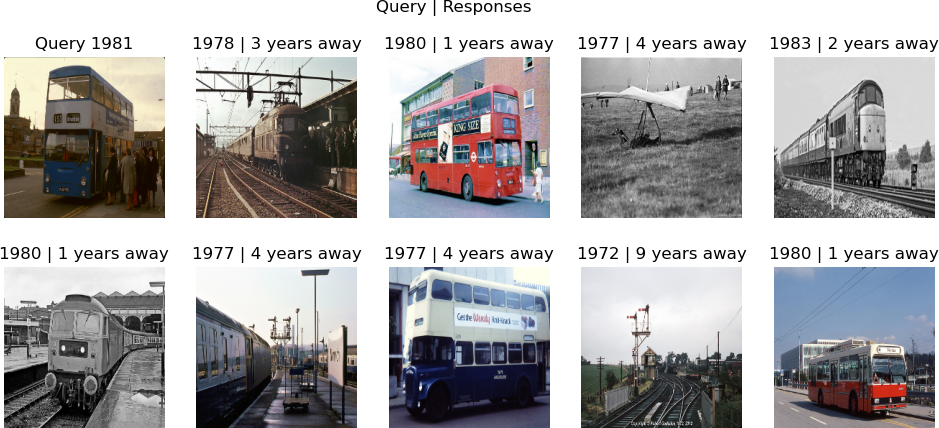}
    \caption{Possible biased retrieval due lack of balance in the training set.}
    \label{fig:BiasedImage}
\end{figure}

In the case of date estimation in the wild \cite{muller2017picture} dataset, we found out some problematic patterns that can lead the retrieval task to certain bias. As there is not a clear balance between categories in the dataset, many classes such as \textit{trains} or \textit{vehicles} may be clustered together regardless the actual dates of the images Figure~\ref{fig:BiasedImage}. However, it is not easy to say when a certain category is most likely to be linked to a certain period of time. For example, selfies are a category way more common nowadays than 20 years ago. Something similar may happen with certain categories that could be biasing the retrieval to a pure object classification task; however we found out many good clues, Figures \ref{fig:DistanceMatrix}, \ref{fig:qualitative},  that indicate our embedding for ranking images is working properly.

Despite further research is needed; smooth-nDCG usage for large-scale image retrieval is a promising novel approach with many practical applications. Since Brown \etal \cite{brown2020smooth} smoothed us the path to minimize the average precision, we propose smooth-nDCG to model more complex retrieval problems where different labels should not be punished equally according to a certain criteria. 
%Additionally, unlike smooth-AP \cite{brown2020smooth}, smooth-NDCG cares about the whole ranking retrived, not only how many good images are, but how are the bad ones distributed along the ranking. This allowed us to retrieval ordered information (such as years), or tree-structured information where different categories can be closer or further depending on how the tree is like. 
As we commented in smooth-nDCG brings a new way of ranking images beyond categorical labels from a neural network approach; then we would like to consider this approach as an application example for larger information retrieval problems with a more complex structure.

\section*{Acknowledgment}

This work has been partially supported by the Spanish projects RTI2018-095645-B-C21, and FCT-19-15244, and the Catalan projects 2017-SGR-1783, the Culture Department of the Generalitat de Catalunya, and the CERCA Program / Generalitat de Catalunya.

%\input{tex/res.tex}


\bibliographystyle{splncs04}
\bibliography{refs}

\end{document}